\documentclass{article}

\usepackage{arxiv}

\usepackage[utf8]{inputenc} 
\usepackage[T1]{fontenc}    
\usepackage{hyperref}       
\usepackage{url}            
\usepackage{booktabs}       
\usepackage{amsfonts}       
\usepackage{nicefrac}       
\usepackage{microtype}      
\usepackage{lipsum}
\usepackage{graphicx}
\usepackage{nccmath}
\usepackage{subcaption}
\usepackage{bm}
\usepackage{hyperref}
\graphicspath{ {./images/} }
\hypersetup{
    colorlinks=true,
    linkcolor=black,
    filecolor=magenta,      
    urlcolor=blue,
    citecolor=blue,
}

\title{Efficient-CapsNet: Capsule Network with Self-Attention Routing}

\author{
 Vittorio Mazzia \\
  Department of Electronics and Telecommunications\\
  Politecnico di Torino\\
  Turin, Italy 10124\\
  \texttt{vittorio.mazzia@polito.it} \\
   \And
 Francesco Salvetti \\
  Department of Electronics and Telecommunications\\
  Politecnico di Torino\\
  Turin, Italy 10124 \\
  \texttt{francesco.salvetti@polito.it} \\
  \And
  Marcello Chiaberge \\
  Department of Electronics and Telecommunications\\
  Politecnico di Torino\\
  Turin, Italy 10124 \\
  \texttt{marcello.chiaberge@polito.it} \\
}

\begin{document}
\maketitle
\begin{abstract}
Deep convolutional neural networks, assisted by architectural design strategies, make extensive use of data augmentation techniques and layers with a high number of feature maps to embed object transformations. That is highly inefficient and for large datasets implies a massive redundancy of features detectors. Even though capsules networks are still in their infancy, they constitute a promising solution to extend current convolutional networks and endow artificial visual perception with a process to encode more efficiently all feature affine transformations. Indeed, a properly working capsule network should theoretically achieve higher results with a considerably lower number of parameters count due to intrinsic capability to generalize to novel viewpoints. Nevertheless, little attention has been given to this relevant aspect. In this paper, we investigate the efficiency of capsule networks and, pushing their capacity to the limits with an extreme architecture with barely 160K parameters, we prove that the proposed architecture is still able to achieve state-of-the-art results on three different datasets with only 2\% of the original CapsNet parameters. Moreover, we replace dynamic routing with a novel non-iterative, highly parallelizable routing algorithm that can easily cope with a reduced number of capsules. Extensive experimentation with other capsule implementations has proved the effectiveness of our methodology and the capability of capsule networks to efficiently embed visual representations more prone to generalization.
\end{abstract}


\section{Introduction}
\begin{figure}
    \centering
    \includegraphics[scale=1.1]{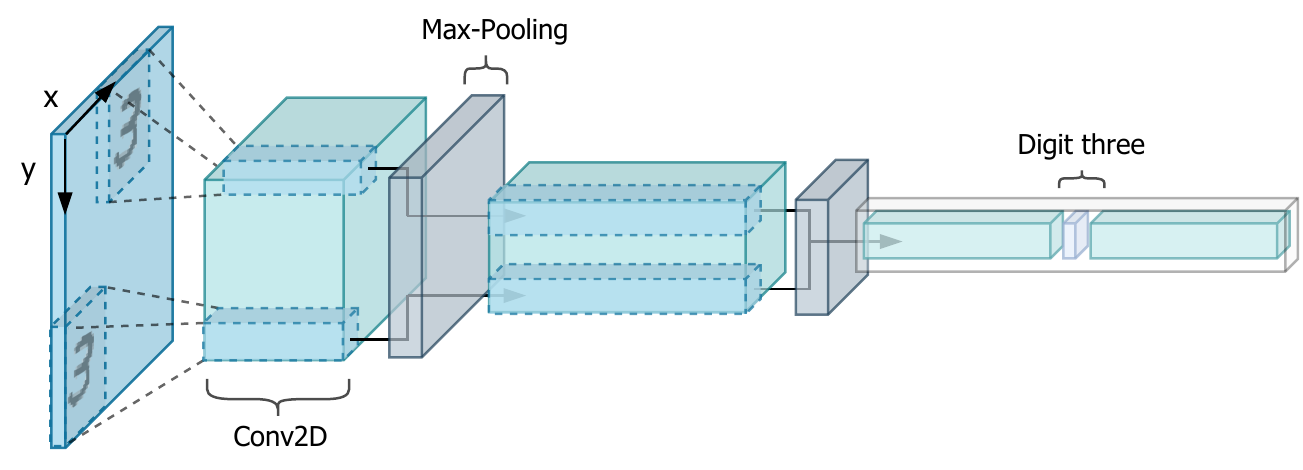}
    \caption{Compressed representation of a simple CNN with max-pooling layers for spatial reduction and two input objects obtained with a plain spatial translation. Max-pooling operations are schematized in such a way that their primitive routing role is highlighted for both digits. Low-level features detected in the earlier stage of the network are progressively routed to common high-level features. So, the model is translation invariant but gradually loses relevant object localization information.}
    \label{fig:translation_invariance}
\end{figure}

In the last decade, convolutional neural networks (CNN) drastically changed artificial visual perception, achieving remarkable results in all core fields of computer vision, from image classification \cite{krizhevsky2017imagenet,he2016deep,hu2018squeeze} to object detection  \cite{redmon2016you,liu2016ssd,mazzia2020real} and instance segmentation \cite{he2017mask}. In contrast to other deep neural architectures, the main characteristic of a CNN is its capability to efficiently replicate the same knowledge at all locations in the spatial dimension of an input image. Indeed, using translated replicas of learned feature detectors, features learned at one spatial location are available at other locations. Local shared connectivity coupled with spatial reduction layers, such as max-pooling, extract local translation-invariant features. So, as shown in Figure \ref{fig:translation_invariance}, object translations in the input space do not affect activations of high-level neurons, because max-pooling layers are able to rout low-level features between the layers. Nevertheless, translation invariance achieved by CNN comes at the expense of losing the precise encoding of objects location. Moreover, CNNs are not invariant to all other affine transformations.

During the years, different techniques have been developed to counterbalance that problem. Most of the adopted common solutions make use of an increased number of feature maps in such a way that the network is endowed with enough feature detectors for all additional transformations. Data augmentation techniques are used to produce the different pose to be learned, and residual connections and normalization techniques allow to enlarge networks filter capacity. However, all those additional mechanisms only partially make up for the intrinsic limitations of CNN, preventing the model from recognising different transformations of the same objects encountered during training. Indeed, CNNs trained on large datasets have a massive redundancy of features detectors and difficulties to scale to thousands of objects with their respective viewpoints.

Hinton et al. \cite{hinton2011transforming} proposed to make neurons cooperate in a new form of unit, dubbed capsules, where individual activations inside them do not represent the presence of a specific feature but different properties of the same entity anymore. In their paper they showed that groups of neurons, if properly trained, are able to produce a whole vector of numbers, explicitly representing the pose of the detected entity as in classical hand-engineered features \cite{lowe1999object}. After six years, Sabour et al. \cite{sabour2017dynamic} presented a first architecture, named CapsNet, that introduced capsules inside a CNN. The major insight of the paper is that viewpoint changes have complicated effects on the pixel space, but simple linear effects on the pose that represents the relationship between an object-part and the whole. In a generic fully-connected or convolutional deep neural network, weights are used to encode feature detectors and neuron activations to represent the presence of a specific feature. So, fixing weights after training, the model is not able to detect simple transformation patterns not encountered during training. On the other hand, they suggested repurposing weights to embed relationships between object features. Indeed, being intrinsic transformation between parts and a whole invariant to the viewpoint, weights are perfectly fitted to represent them efficiently, and they should be automatically capable of generalizing to novel viewpoints. Moreover, we do not want anymore to achieve activations invariant to transformations, but groups of neurons working in synergy to represent different properties of the same entity. Capsules are vector representations of features, and they are equivariant to viewpoint transformation. So, each capsule not only represents a specific type of entity but also dynamically describes how the entity is instantiated. Finally, the working principle of traditional networks, in which a scalar unit is activated based on the matching score with learned feature detectors, is dropped altogether favouring a much more robust mechanism. Indeed, with viewpoint invariant transformations encoded in the weights, we can make capsules predict the whole that they should be part of. So, we can consider predictions accordance of low-level capsules to activate high-level capsules. That requires a process to measure their agreement and route capsules to their best match parent. Originally, dynamic routing was proposed as the first routing-by-agreement mechanism.  Exploiting groups of neuron activations to make predictions and assess their reciprocal agreement is a much more effective way to capture covariance and should lead to models with a considerably reduced number of parameters and far better generalization capabilities.

Nevertheless, little attention has been given to the efficiency aspect of capsule networks and their intrinsic capability to represent knowledge object transformations better. Indeed, all model solutions presented so far account for a large number of parameters that inevitably hide the intrinsic generalization capability that capsules should provide. In this paper, we propose Efficient-CapsNet, an extreme architecture with barely 160K parameters and a 85\% TOPs improvement upon the original CapsNet model that is perfectly capable of achieving state-of-the-art results on three distinct datasets, maintaining all important aspects of capsule networks. With extensive experimentation with traditional CNNs and other capsule implementations, we proved the effectiveness of our methodology and the important contribution lead by capsules inside a network. Moreover, we propose a novel non-iterative, routing algorithm that can easily cope with a reduced number of capsules exploiting a self-attention mechanism. Indeed, attention, as also max-pooling layers, can be seen as a way to route information inside a network. Our proposed solution  exploits similarities between low-level capsules to cluster and routs them to more promising high-level capsules. Overall, the main contribution of our work lies in:

\begin{itemize}
    \item Deep investigation of the generalization power of networks based on capsules, drastically reducing the number of trainable parameters compared to previous literature research studies.
    \item The Conceptualization and development of an efficient, highly replicable, deep learning neural network based on capsules able to reach state-of-the-art results on three distinct datasets.
    \item The introduction of a novel non-iterative, highly parallelizable routing algorithm that exploits a self-attention mechanism to route a reduced number of capsules efficiently.
\end{itemize}
All of our training and testing code are open source and publicly available\footnote[1]{https://github.com/EscVM/Efficient-CapsNet}.
The remainder of this paper is structured as follows. Section
II covers the related work on capsule networks, their developments in the latest years and practical applications. Section III provides a comprehensive overview of the methodology, network architecture and its routing algorithm. Section IV discusses the experimentation and results with three datasets, MNIST, smallNorb and MultiMNIST. Moreover, it provides an introspect analysis of the inner operation of capsules inside a network. Finally, section V draws some conclusions and future directions.

\section{Related Works}
\begin{figure}[t]
    \centering
    \includegraphics[scale=1]{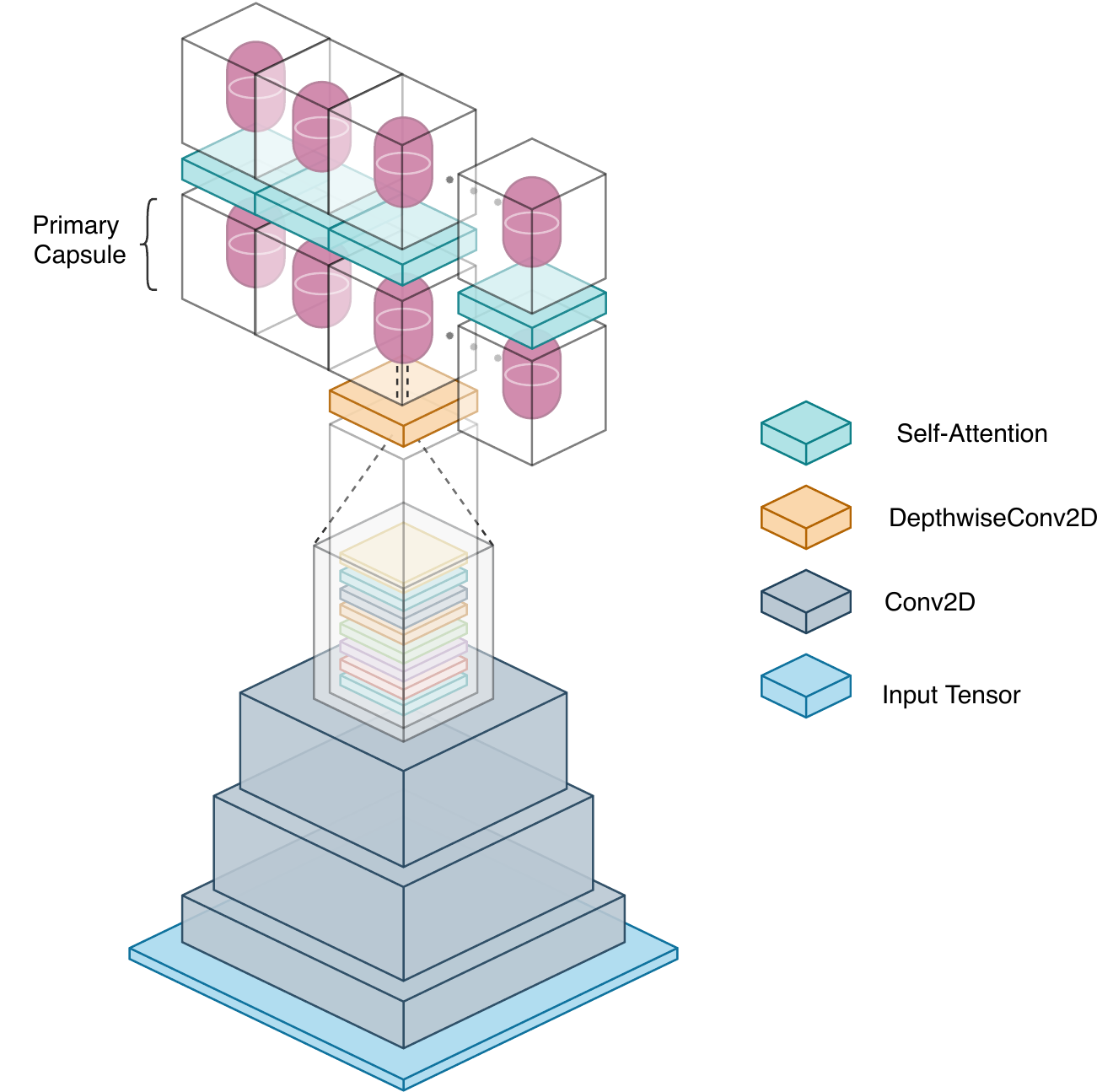}
    \caption{Schematic representation of the overall architecture of Efficient-CapsNet. Primary capsules make use of depthwise separable convolution to create a vectorial representation of the features they represent. On the other hand, the first stack of convolutional layers maps the input tensor onto a higher-dimensional space, facilitating capsules creation.}
    \label{fig:network}
\end{figure}
As already devised in the introduction to this paper, introducing a vectorial organization of neurons to encapsulate both probability and instantiation parameters of a detected feature was first proposed by Hinton et al. \cite{hinton2011transforming} introducing the new concept of capsules. Sabour et al. \cite{sabour2017dynamic} proposed the first CNN able to incorporate two layers of capsules, called CapsNet, and introduced the routing-by-agreement concept, with their dynamic routing. Several researchers have then investigated the routing process, proposing alternative ways to measure accordance between low-lever capsules in activating high-level ones.

Xi et al. \cite{xi2017capsule} proposed a variant to the squash activation function used in the original CapsNet. Wang et al. \cite{wang2018optimization} gave a formal description of the original dynamic routing as an optimization problem that minimizes clustering loss and proposes a slightly modified version. Lenssen et al. \cite{lenssen2018group} proposed group capsule networks, claiming they preserve equivariance for the output pose and invariance for activations. The same authors of the original CapsNet adapted the Expectation-Maximization algorithm to cluster similar votes, and route predictions \cite{hinton2018matrix}. Spectral capsule network \cite{bahadori2018spectral} was based on this last work, and modified routing basing it on Singular Value Decomposition of votes from the previous layers. Ribeiro et al. \cite{ribeiro2020capsule} proposed a routing derived from Variational Bayes for fitting a gaussian mixture model. Gu et al. \cite{gu2020improving} focused on making capsule networks 
robust to affine transformations by sharing transformation matrices between all low-level capsules and each high-level ones. Paik et al. \cite{paik2019capsule} put in discussion the effectiveness of the routing algorithm presented so far, claiming that better results can be obtained with no routing at all. On the other hand, Venkataraman et al. \cite{venkatraman2020learning} proved that routing-by-agreement mechanism is essential to ensure compositional structures of capsule-based networks. Byerly et al. \cite{byerly2020branching}, instead, proposed a new architecture based on a variation of the original capsule idea, named Homogeneous Filter Capsules, and with no routing between layers.

The attention mechanism allows to dynamically give more importance to particular features that are considered more relevant for the problem under analysis. Such an idea gained great popularity in a number of Deep Learning applications and have been implemented in natural language processing  \cite{bahdanau2014neural,vaswani2017attention} or computer vision  \cite{jaderberg2015spatial,xu2015show,hu2018squeeze,woo2018cbam, salvetti2020multi}. Choi et al. \cite{choi2019attention} applied the attention mechanism to capsule routing with a feed-forward operation with no iterations. However, they selected low-level capsules by multiplying their activations to a parameter vector learnt with backpropagation, and they did not measure agreement. In this way, the original idea of routing-by-agreement is drastically modified. Tsai et al. \cite{tsai2020capsules} slightly changed the original dynamic routing to compute the agreement between a pose of a high-level capsule and the votes of the low-level capsules by an inverted dot-product mechanism. They proposed a concurrent iterative routing instead of a sequential one, performing the routing procedure simultaneously on all the capsule layer. Huang et al. \cite{huang2020capsnet} proposed a dual attention mechanism by adapting the squeeze-and-excitation block \cite{hu2018squeeze} to both Primary and Digit Caps, together with a change in the squash activation function. Peng et al. \cite{peng2020bg} applied capsules in addition with a self-attention based backbone for an entity relation task in natural language processing. However, both these last two works used attention mechanisms as part of the computational graph of the proposed networks, without modifying the original dynamic routing proposed by Sabour et al. \cite{sabour2017dynamic}. In this sense, our approach strongly differs from theirs since we first propose self-attention as a substitute routing algorithm between capsules. Capsule-based networks have also been recently used for a variety of applications. For example, they have been applied for natural language processing \cite{peng2020bg,mcintosh2018multi,zhang2018attention,du2019novel}, with GANs for image generation \cite{jaiswal2018capsulegan}, computer vision \cite{duarte2018videocapsulenet,lalonde2018capsules,nguyen2019capsule} or medicine \cite{mobiny2019automated,kruthika2019cbir}.

\section{Methods}
\subsection{Efficient-CapsNet}
The overall architecture of Efficient-CapsNet is depicted in Figure \ref{fig:network}. As a high-level description, the network can be broadly divided into three different parts in which the first two are the main instruments of the primary capsule layer to interact with the input space. Indeed, each capsule exploits the below convolutional layer filters to convert pixel intensities into a vectorial representation of the feature it acts for. So, the activities of neurons within an active capsule embody the various properties of the entity it learnt to represent during the training process. As stated in Sabour et al. \cite{sabour2017dynamic}, these properties can include many different types of instantiation parameter such as pose, texture, deformation, and among those the existence of the feature itself. In our implementation, the length of each vector is used to represent the probability that the entity represented by a capsule is present. That is compatible with our self-attention routing algorithm that does not require any sensible objective function minimization. Moreover, it makes biological sense as it does not use large activities to represent absent entities. Finally, the last part of the network operates under the self-attention algorithm to rout low-level capsules to the whole they represent. 

More formally, in the case of a single instance $(i)$, the model takes as input an image that can be represented as a tensor $\textbf{X}$ with a shape $H \times W \times F$ where $H$, $W$ and $C$ are the height, width, and channels/features of the single input image. Before entering the primary caps layer, we extract local features from the input image $\textbf{X}$ by means of a set of convolutional and Batch Normalization layers \cite{ioffe2015batch}. Each output of a convolution layer $l$ is constituted by a convolutional operation with a certain kernel dimension $k$, number of feature maps $f$, stride $s=1$ and ReLU as activation function:
\begin{ceqn}
\begin{gather}
    \mathrm{F}^{l+1}(\textbf{X}^{l}) = \mathrm{ReLU}\left(Conv_{k \times k}(\textbf{X}^{l})\right)
\end{gather}
\end{ceqn}

\begin{figure}
    \centering
    \includegraphics[scale=1]{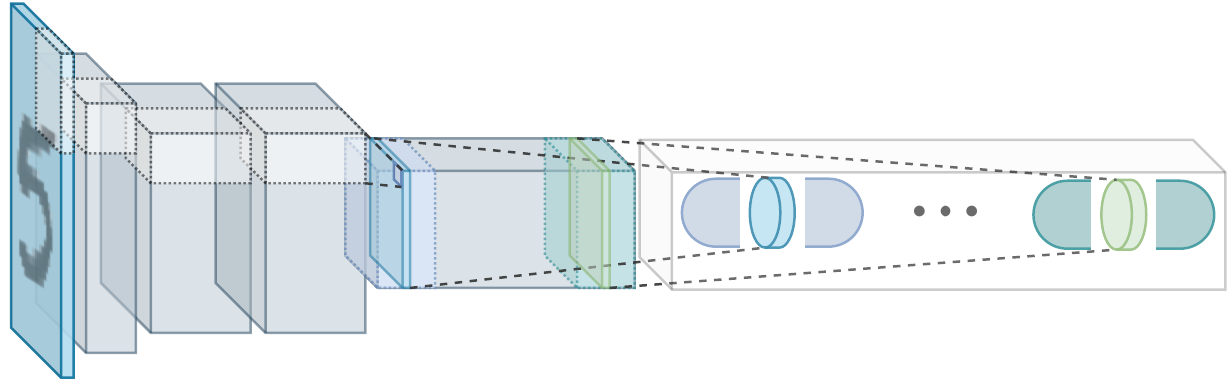}
    \caption{The first part of the network can be modelled as single-function $H_{Conv}$ that maps the input image onto a higher-dimensional space. Then, the primary capsule layer $\textbf{\textit{S}}^{l}_{n,d}$ is obtained with a depthwise separable convolution that greatly reduces the number of parameters needed for the capsules creation.}
    \label{fig:creation_capsules}
\end{figure}

Overall, the first convolutional part of the network can be modelled as a single function $H_{Conv}$ that maps the input image onto a higher dimensional space that facilitates the capsule creation. On the other hand, the second part of the network is the main instrument used by primary capsules to create a vectorial representation of the features they represent. As depicted in Figure \ref{fig:creation_capsules}, it is a depthwise separable convolution with linear activation that performs just the first step of a depthwise spatial convolution operation, acting separately on each channel. Moreover, imposing a kernel dimension $k \times k$ and a number of filters $f$ equal to the output dimensions $H \times W$ and $F$ of the $H_{Conv}$ function, it is possible to obtain the primary capsule layer 
$\textbf{\textit{S}}^{l}_{n,d}$ where $n^l$ and $d^l$ are the number of primary capsules and their individual dimension of the $l-th$ layer, respectively.
The depthwise separable convolution is an efficient operation that greatly simplifies and reduces the number of parameters required for the capsule creation process. We leave it to discriminative learning to make good use of its filters to smartly extract all capsule properties.

After that operation, location information is not anymore "place-coded" but "rate-coded" in the properties of the capsules. So, the base element of the network is not anymore a single neuron but a vector-output capsule. Indeed, the first operation applied to the primary capsule layer is a capsule-wise activation function. In order to encode the probability that a certain entity exists with the length of vectors and let active capsules make predictions for the instantiation parameters of higher-level capsules, two important properties should be satisfied by the activation function; it should preserve a vector orientation and maintain the length between zero and one. Efficient-CapsNet makes use of a variant of the original activation function, dubbed squash operation:
\begin{ceqn}
\begin{gather}
    \textrm{squash}(\textbf{\textit{s}}^{l}_{n})=\left(1 - \frac{1}{e^{||\textbf{\textit{s}}^{l}_{n}||}}\right)\frac{\textbf{\textit{s}}^{l}_{n}}{||\textbf{\textit{s}}^{l}_{n}||}
    \label{eq:squash_variant}
\end{gather}
\end{ceqn}
where we refer to a single capsule as $\textbf{\textit{s}}^{l}_{n}$, which are the individual entries $n^l$ of $\textbf{\textit{S}}^{l}_{(n,:)}$\footnote[1]{$\textbf{\textit{s}}^{l}_{n_0}:=\{\textbf{\textit{S}}^{l}_{n,d}|n^l=n_0^l\}$} with $\textbf{\textit{s}}^{l}_{n} \in \mathbb{R}^{d^l}$. The capsule-wise squash function of Eq. (\ref{eq:squash_variant}), satisfies the required two properties and is much more sensitive to small changes near zero, providing a boost to the gradient during the training phase \cite{xi2017capsule}. So, after the squash activation we obtain a new matrix $\textbf{\textit{U}}^{l}_{n,d}$ with all $n^l$ entries $\textbf{\textit{u}}^{l}_{n}$ with the same dimensionality and properties of $\textbf{\textit{s}}^{l}_{n}$, but with a length "squashed" between zero and one. Indeed the non-linearity ensure that short vectors get shrunk to almost zero length and long vectors get shrunk to a length slightly below one.

\begin{figure}[t]
    \centering
    \includegraphics[scale=0.78]{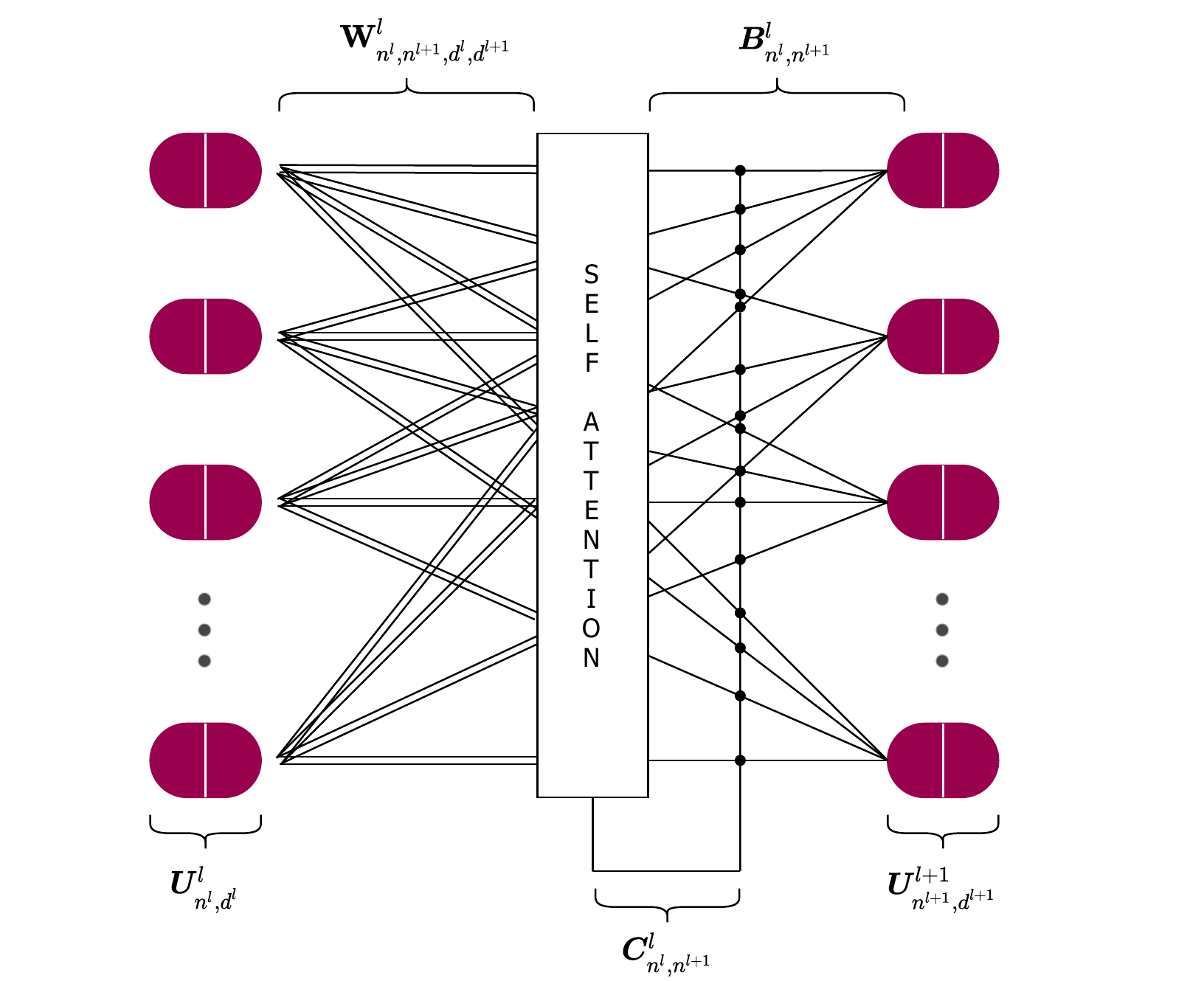}
    \caption{Capsules of the layer $l-th$ make predictions of the whole they could be part of. All predictions obtained with the weight tensor $\textbf{W}^{l}_{n^{l},n^{l+1},d^{l},d^{l+1}}$ are collected in $\hat{\textbf{U}}_{n^l,n^{l+1},d^{l+1}}^{l}$ that is subsequently used in conjunction with the priors $\textbf{\textit{B}}^l_{n^l,n^{l+1}}$ and coupling coefficients $\textbf{\textit{C}}^{l}_{n^l,n^{l+1}}$ matrices to obtain all capsules $\textbf{\textit{s}}^{l+1}_{n}$ of layer $l+1$.}
    \label{fig:routing_capsules}
\end{figure}
\subsection*{Self-Attention routing}
In order to rout active capsules to the whole they belong, we make use of our self-attention routing algorithm. As shown in Figure \ref{fig:routing_capsules}, despite the additional dimension, the overall architecture is very similar to a fully-connected network with an additional branch brought by the self-attention algorithm. Indeed, the total input of a capsule  in the above layer, $\textbf{\textit{s}}^{l+1}_{n}$, is a weighted sum over all "prediction vectors" from the capsules $\textbf{\textit{u}}^{l}_{n}$ in the layer below. That is produced by a matrix multiplication of each capsule, $\textbf{\textit{u}}^{l}_{n}$, belonging to $\textbf{\textit{U}}^{l}_{n,d}$, for a weight matrix. Intuitively, the whole tensor $\textbf{W}^{l}_{n^{l},n^{l+1},d^{l},d^{l+1}}$ that contains all weight matrices, embeds all affine transformation between capsule of two adjacent layers. So, each capsule of the layer $l$, in order to make its projections for the layer above, follows Eq. \ref{eq:prediction}
\begin{ceqn}
\begin{gather}
    \hat{\textbf{U}}_{(n^l,n^{l+1},:)}^{l}=\textbf{\textit{u}}_{n}^{\textrm{T}l}\times\textbf{W}_{(n^l,n^{l+1},:,:)}^{l}
    \label{eq:prediction}
\end{gather}
\end{ceqn}
 where $\hat{\textbf{U}}_{n^l,n^{l+1},d^{l+1}}^{l}$ contains all predictions of $l-th$ capsules. Indeed, each $n^l$ capsule, by means of the weight matrix, predicts the properties of all $n^{l+1}$ capsules. Indeed, capsules of the above layer, $\textbf{\textit{s}}^{l+1}_{n}$, can be computed with Eq. \ref{eq:capsule_output}
\begin{ceqn}
\begin{gather}
    \textbf{\textit{s}}^{l+1}_{n} = \hat{\textbf{U}}_{(:,n^{l+1},:)}^{\textrm{T}l} \times \left(\textbf{\textit{C}}^{l}_{(:,n^{l+1})} + \textbf{\textit{B}}^l_{(:,n^{l+1})}\right)
    \label{eq:capsule_output}
\end{gather}
\end{ceqn}
where $\textbf{\textit{B}}^l_{n^l,n^{l+1}}$ is the log priors matrix containing all weights discriminatively learnt at the same time as all the other weights. On the other hand, $\textbf{\textit{C}}^{l}_{n^l,n^{l+1}}$ is the matrix containing all coupling coefficients produced by the self-attention algorithm. So, the priors help to create biases towards more linked capsules and the self-attention routing dynamically assigns detected shapes to the whole they represent in the specific $(i)$ instance taken into account. The coupling coefficients are computed starting from the self-attention tensor $\textbf{A}^l_{n^l,n^l,n^{l+1}}$ using Eq. \ref{eq:self_attention}
\begin{ceqn}
\begin{gather}
    \textbf{A}^l_{(:,:,n^{l+1})} = \frac{\hat{\textbf{U}}_{(:,n^{l+1},:)}^{l} \times \hat{\textbf{U}}_{(:,n^{l+1},:)}^{\textrm{T}l}}{\sqrt{d^l}}
    \label{eq:self_attention}
\end{gather}
\end{ceqn}
which contains a symmetric matrix $\textbf{A}^l_{:,:,n^{l+1}}$ for each capsule $n^{l+1}$ of the layer above. The term $\sqrt{d^l}$ stabilizes training and helps maintaining a balance between coupling coefficients and log priors. Each self-attention matrix contains the score agreement for each combination of the $n^l$ capsules predictions, and so, they can be used to compute all coupling coefficients. In particular, Eq.\ref{eq:softmax_routing} is used to compute the final coefficients that can be used in Eq. \ref{eq:capsule_output} to obtain all capsules $\textbf{\textit{S}}^{l+1}_{n,d}$ of the layer $l+1$.
\begin{ceqn}
\begin{gather}
    \textbf{\textit{C}}^{l}_{(:,n^{l+1})} = \frac{exp\left(\sum_{n^l}\textbf{A}^l_{(:,n^l,n^{l+1})}\right)}{\sum_{n^{l+1}}exp\left(\sum_{n^l}\textbf{A}^l_{(:,n^l,n^{l+1})}\right)}
    \label{eq:softmax_routing}
\end{gather}
\end{ceqn}
So, the coupling coefficients between a capsule of layer $l$ and all the capsules in the layer above, $l+1$, sum to one. Successively, initial log prior probabilities are add to the coupling coefficients to obtain the final routing weights. The procedure remains unchanged in presence of multiple capsule layers, stacked on top of each other in order to create a deeper hierarchy. 
\subsection{Margin Loss and reconstruction regularizer}
The output layer is not anymore represented by a scalar, but by a vector as well. Indeed, a capsule of the final layer does not only represent the probability that a certain object class exists, but also all its properties extracted from its individual parts. The length of the instantiation vector is used to represent the probability that a capsule's entity exists. Its length should be close to one if and only if the entity it represents is the only one present in the image. So, to allow multiple-class, we compute Eq. \ref{eq:margin_loss} for each class represented by a capsule $n^L$ of the last layer $L$:
\begin{ceqn}
\begin{gather}
    \mathcal{L}_{n^L} = T_{n^L}\textrm{max}\left(0,m^{+} - ||\textbf{\textit{u}}^L_n||\right)^{2} + \lambda\left(1-T_{n^L}\right)\textrm{max}\left(0,||\textbf{\textit{u}}^L_n|| - m^{-}\right)^{2}
    \label{eq:margin_loss}
\end{gather}
\end{ceqn}
where $T_{n^L}$ is equal to one if the class $n^L$ is present and $m^{+}$, $m^{-}$ and $\lambda$ are hyperparameters to be tuned. Then, the separate margin loss $\mathcal{L}_{n^L}$ are summed to compute the final score during the training phase.

Finally, we adopt the reconstruction regularizer as in \cite{sabour2017dynamic} to encourage all final capsules to encode robust and meaningful properties. So, the output capsules $\{\textbf{\textit{u}}^L_n\}_{n=1,...,N}$ are fed to the reconstruction decoder and the mean of L2 loss between an input image and the decoder output is added to the marginal loss scaled by a factor $r$.  

\section{Results}
We aim to simply demonstrate that a properly working capsule network should achieve higher results with a considerably lower number of parameters due to its intrinsic capability to embed information better and efficiently. In this section, we test the proposed methodology in an experimental context, assessing its generalization capabilities and efficiency respect to traditional convolutional neural networks and similar works present in literature. 
On this purpose, we test our proposed methodology with three of the most used dataset for capsule-based networks assessment: MNIST, smallNORB and MultiMNIST. On all datasets, we demonstrate a remarkable difference with traditional solutions and comparable accuracy levels with similar methodologies but with a fraction of the trainable parameters in most cases. All experimentation clearly shows that a capsule network is capable to achieve higher results with a considerably lower number of parameters count. Moreover, we show how a simple ensemble of a few instances of Efficient-CapsNet can easily establish state-of-the-art results in all the three datasets. Finally, using principal component analysis, we give an introspect to the inner representations of the network and its capability to encode visual information.

\subsection{Experimental settings}
\begin{table}[b]
\centering
\begin{tabular}{lccc}
\toprule
Method                         & Parameters {[}K{]} & OPS$|_{1batch}$ {[}G{]} & Improvement$|_{1batch}$ (\%) \\ \hline
CapsNet \cite{sabour2017dynamic}        & 6800 &        0.401        & 84.96     \\
AR CapsNet \cite{choi2019attention}     & 5310 &        0.098        & 38.66   \\ 
Matrix-CapsNet with EM routing \cite{hinton2018matrix} & 310  & 0.086               & 29.56                   \\ 
Efficient-CapsNet      & 161 &         0.06                & -                       \\ \bottomrule
\end{tabular}
\caption{Comparison of the computational cost in terms of necessary operations between Efficient-CapsNet and other similar methodologies present in literature. Efficient-CapsNet, besides having a reduced number of trainable parameters, is much more efficient.}
\label{table:computation_cost}
\end{table}
In all experiments, in order to map input samples onto an higher dimensional space, we adopt four convolutional layers with $k=5$ for the first convolution and $k=3$ for all others. On the other hand, $f$ is equal to 32, 64, 64 and 128, respectively. ReLU is used in all layers, but leaky-ReLU is a valuable alternative. As previously discussed, the number of capsules depend by the number of feature maps, $f$, of the last convolutional layer. Indeed, the depthwise separable operation has a kernel dimension $k \times k$ equal to the output dimension $H \times W $ of the $H_{Conv}$ function and a number of filters $f$ equal to its filter dimension $F$. The first layer of primary capsules, $\textbf{\textit{S}}^1_{n,d}$, has $n^1=16$ capsules with a dimension $d^{1}$ of 8.
Multiple fully-connected capsule layers can be added to increase the capacity of the network. However, we adopt only two layers of capsules due to the relative simplicity of the dataset investigated. Finally, the output layer of the network has a number of capsules $n^L$ equal to the classes of the specific dataset taken into account. Since that higher-level capsules represent more complex entities with more degrees of freedom, their capsules dimensionality increases. 

All loss parameters are obtained by CapsNet \cite{sabour2017dynamic} training. So, for all experimentation $m^{+}$, $m^{-}$ and $\lambda$ are set to 0.9, 0.1 and 0.5, respectively. Moreover, the scaling factor $r$ for the reconstruction regularizer is set to 0.392. Indeed, since we use the mean of L2 loss, while CapsNet uses the sum of L2 loss, $0.392=0.0005 * 784$. All experimentations are carried out on a workstation with an Nvidia RTX2080 GPGPU with 8GB of memory and 64GB of DDR4 SDRAM. We use the TensorFlow 2.x framework with CUDA 11. All result statistics are obtained with a mean of 30 trials.
 
In Table \ref{table:computation_cost} is presented a comparison between the architecture of Efficient-CapsNet and other similar methodologies. Our model has a much lower number of parameters count, and it is much more efficient in terms of operations required. So, it can clearly highlight the generalization capability of capsules with respect to traditional CNN.

\subsection{MNIST results}
\begin{figure}
    \centering
    \includegraphics[scale=0.45]{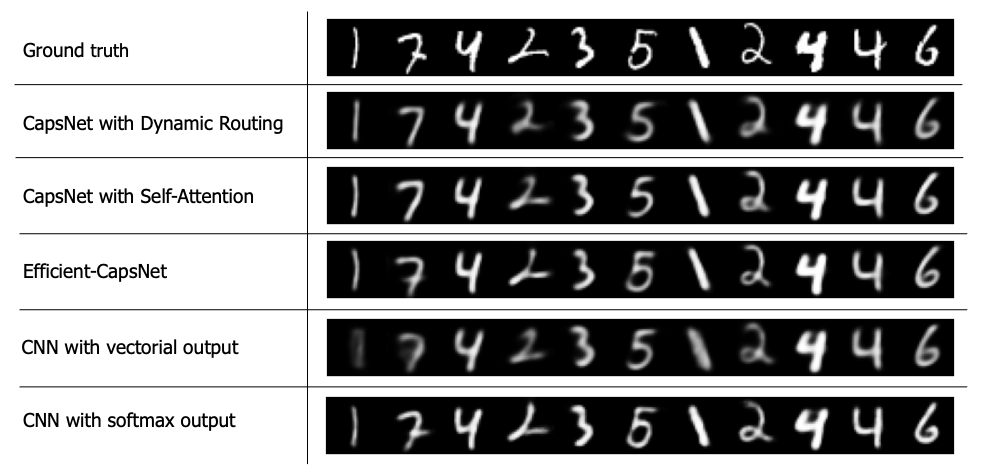}
    \caption{Digit reconstruction with different tested methodologies. Even with different architecture strategies and training objectives, all networks are able to embed different properties of the input digits keeping only important details.}
    \label{fig:generation}
\end{figure}

The MNIST dataset \cite{lecun1998mnist} is composed of 70000, $28 \times 28$, images divided in 60000 and 10000 for training and testing, respectively. We adopt the same data augmentation proposed in Byerly et al. \cite{byerly2020branching}. The reconstruction network is a simple fully-connected network with two hidden layers with 512 and 1024 neurons. We test our methodology and compare it with different models and two custom CNN baseline. In particular, our baseline is identical to Sabour et al. \cite{sabour2017dynamic} with the exception of a reduced number of feature maps and layers, in order to keep the number of parameters as close as possible to Efficient-CapsNet. On the other hand, "Base-CapsNet" is a CNN but with a vectorial output as in a capsule-based network. So, it is also trained with the marginal loss function. That is specifically devised to assess the role of the reconstruction network and its impact on the overall accuracy. Our networks are trained for 100 epochs, batch size of 16, Adam \cite{kingma2014adam} optimizer and an initial learning rate of $\eta=5e-4$ with exponential decay 0.98. All hyperparameters are selected with a small percentage of validation data.

\begin{table}[h!]
\centering
\begin{tabular}{lccc}
\toprule
Method            & Reconstruction & Parameters {[}K{]} & MNIST {[}\%{]}                  \\ \hline
Our Baseline      & no              & 173                & 0.48                        \\ 
Base-CapsNet      & no              & 183                & 0.54                        \\ 
Our Baseline      & yes              & 173                & 0.4                         \\ 
Base-CapsNet       & yes              & 183                & 0.39                        \\ 
Efficient-CapsNet & yes              & 161                & $0.26_{\pm 0.0002}$              \\ \hline
Baseline \cite{sabour2017dynamic}          & no              & 35400              & 0.39                        \\ 
CapsNet \cite{sabour2017dynamic}           & yes              & 6800               & $0.25_{\pm 0.005}$ ($0.36_{\pm 0.04}$)* \\
Matrix-CapsNet with EM routing \cite{hinton2018matrix}  & no              & 310               & 0.44                        \\ 
DA-CapsNet \cite{huang2020capsnet}        & yes              & 7000               & 0.47                        \\ 
AR CapsNet \cite{choi2019attention} & yes & 5310 & 0.54 \\
HFCs \cite{byerly2020branching}              & no              & 1514               & $0.25_{\pm 0.0002}$              \\ \bottomrule
\end{tabular}
\caption{Test error (\%) on the MNIST classification task. All methodologies are reported with their number of parameters and the presence of the reconstruction regularizer during the training phase. * indicates the results from our experiments.}
\label{tab:mnist_results}
\end{table}

In Table \ref{tab:mnist_results} are reported parameters and test errors of the different tested architectures.  It is evident the gap between all baseline CNNs and all other capsule-based networks. Moreover, even if Efficient-CapsNet has barely 161K parameters, it is comparable with all other methodologies present in the literature so far. It achieves a mean accuracy of 0.9974 with a min value of 0.9971 and a max one of 0.9978. Finally, a network with a vectorial output receives a significant boost in performance using the reconstruction regularizer. In Figure \ref{fig:generation} are presented some images generated by the reconstruction networks of the different tested methodologies. It also worth to notice that, even in the presence of an adaptive gradient descent method, Efficient-CapsNet does not overfit the training set but register a similar accuracy with the test set after the training.
\begin{table}[t]
\centering
\begin{tabular}{lcc}
\toprule
Method                                                  & Year & Test Error {[}\%{]} \\ \hline
Multi-Column Deep Neural Networks for Image Classification \cite{ciregan2012multi} & 2012 & 0.23 \\
Regularization of Neural Networks using DropConnect \cite{wan2013regularization}     & 2013 & 0.21       \\ 
RMDL:Random Multimodel Deep Learning for Classification \cite{kowsari2018rmdl} & 2018 & 0.18       \\ 
Base-Branching \& Merging CNNw/HFCs \cite{byerly2020branching}                    & 2020 & 0.16      \\
Efficient-CapsNet                                       & 2021 & 0.16       \\ \bottomrule
\end{tabular}
\caption{Test error (\%) on the MNIST classification task of state-of-the-art methodologies based on ensemble over the years.}
\label{tab:MNIST_ensemble}
\end{table}

\begin{figure}[b]
    \centering
    \includegraphics[scale=0.69]{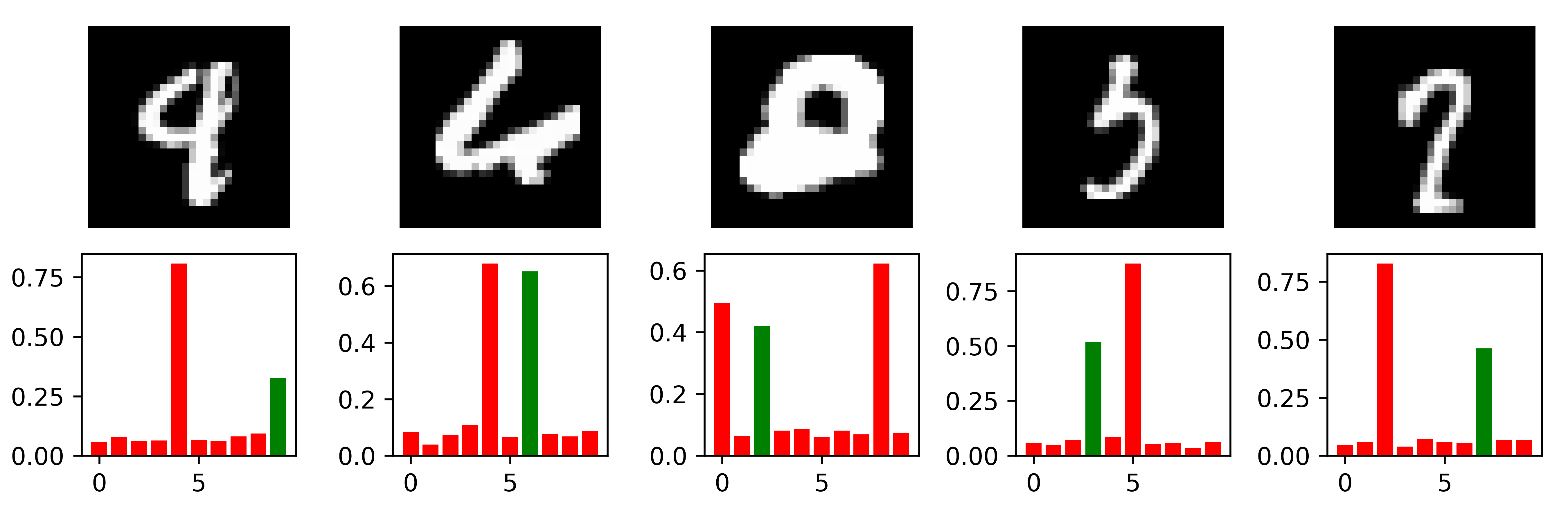}
    \caption{Example of Efficient-CapsNet misclassified digits. Green bars represent correct labels and their high the corresponding capsule length. The ambiguity of these remaining questionable examples is reflected in the uncertainty of the network predictions.}
    \label{fig:uncertainty_digits}
\end{figure}

As previously stated, we also demonstrate that a simple ensemble of Efficient-CapsNet models can easily establish a state-of-the-art result. Indeed, we exploit the 30 trained networks for test score statistics to produce an ensemble prediction. In particular, we average all network predictions with an accuracy greater than 0.9973, obtaining a final test error of 0.16. In Table \ref{tab:MNIST_ensemble} are summarized results of top MNIST leaderboard methodologies. The considerable gap between the mean single network test score, 0.26, and the ensemble one, 0.16, is due to the uncertainty on predictions of all remaining digits. Indeed, Efficient-CapsNet predicts the output class using the length of its output vector. So, unlike the exclusive softmax function, most of the ambiguous digits are reflected in the uncertainty of the network outputs. The ensemble simply steers predictions on the most probable answer. That is a clear sign of the strong knowledge of the dataset encapsulated by the network during the training. Indeed, analyzing the misclassified digits and their prediction scores in the case of a single model clarifies the correctness of its answers despite the given labels. As shown in Figure \ref{fig:uncertainty_digits}, misclassified examples are ambiguous and classifying them correctly is only a matter of pure luck. In our opinion, it is for this reason that networks capable of achieving Efficient-CapsNet level of accuracy have modelled every important aspect of the MNIST dataset and further improvements in the test score have no significant meaning.

\subsection{smallNORB results}
\begin{table}[h]
\centering
\begin{tabular}{lccc}
\toprule
Method                         & Reconstruction & Parameters {[}K{]} & smallNORB {[}\%{]} \\ \hline
Our Baseline                   & no             & 198                & 5.9            \\ 
Base-CapsNet                   & no             & 167                & 4.58           \\ 
Our Baseline                   & yes            & 198                & 4.59           \\ 
Base-CapsNet                   & yes            & 167                & 4.33           \\ 
Efficient-CapsNet              & yes            & 151                & $2.54_{\pm 0.003}$  \\ \hline
Baseline \cite{hinton2018matrix}                      & no             & 4200               & 5.2            \\ 
Matrix-CapsNet with EM routing \cite{hinton2018matrix} & no             & 310                & 1.8 ($4.4_{\pm 0.004}$)*          \\ 
CapsNet \cite{sabour2017dynamic}                        & yes            & 6800               & 3.77           \\ 
VB-Routig \cite{ribeiro2020capsule}                      & yes            & 310                & $1.6_{\pm 0.06}$    \\ \bottomrule
\end{tabular}
\caption{Test error (\%) on the smallNORB classification task. All methodologies are reported with their number of parameters andthe presence of the reconstruction regularizer during the training phase. * indicates the results from our experiments.}
\label{tab:results_smallnorb}
\end{table}
The dataset smallNORB is a collection of 48600 stereo, grayscale images ($96 \times 96 \times 2$), representing 50 toys belonging to 5 generic categories: human, airplanes, trucks, cars and four-legged animals. Each toy was photographed by two cameras under 6 lighting conditions, 9 elevations, and 18 azimuths. The dataset is split in half; 5 instances of each category for the training and the remaining ones for the testing.

Efficient-CapsNet has the same structure described in the "MNIST results" section with the only exception of Instance Normalization \cite{ulyanov2016instance} in place of Batch Normalization layers. That greatly helps the network to deal with different lighting conditions and make the network training as independent as possible of the contrast and brightness differences among the input images.
On the other hand, we follow the same data augmentation and pre-processing proposed in Hinton et. al \cite{hinton2018matrix} with the only exception of the input dimension: we scale the original images to $64 \times 64$ using patches of $48 \times 48$. We train for 200 epochs, with a batch size of 16, Adam optimizer and an initial learning rate of $\eta=5e-4$ with exponential decay of 0.99.

In Table \ref{tab:results_smallnorb} are summarized the results of the baseline networks, Efficient-CapsNet and some capsule-based methodologies present in literature. As for the MNIST dataset, also for smallNORB is evident the gap between classical CNN and capsule-based networks. Moreover, again our methodology has comparable results with all other similar methodologies but with half of the parameters. It achieves a mean accuracy of 0.974 with a min value of 0.97 and a max one of 0.983. Finally, as before we exploit the 30 networks, trained for statistical evidence, to produce an ensemble prediction. We select only the two networks with the lowest test error, and we adopt for both a 40 patch prediction \cite{hinton2018matrix} before averaging their results. We obtain a test accuracy of 1.23, setting a new state-of-the-art result for this dataset.

\subsection{MultiMNIST results}
The MultiMNIST dataset has been proposed by Sabour et al. \cite{sabour2017dynamic} and is based on the superposition of couples of shifted digits from the MNIST dataset. Each original image is first padded to a $36 \times 36$ pixels dimension. A MultiMNIST sample is generated by overlaying two padded digits, which shifts up to 4 pixels in both dimensions, resulting in an average 80\% overlap. The only condition to be met is that the two digits are of different classes. In the labels, both indexes corresponding to the two classes are set to 1. In this way, the network aim is to detect both the digits concurrently. During training, the output capsules corresponding to the target classes are selected one at a time and used to reconstruct the two input images, while during testing we select the two most active capsules, i.e. the longest. Ideally, the network should be able to segment the two digits that have generated the MultiMNIST sample and independently reconstruct them. During training, for each epoch, we randomly generate 10 MultiMNIST images for each original MNIST example. We train the model 5 times independently for about 100 epochs, with a batch size of 64, Adam optimizer and an initial learning rate of $\eta=5e-4$ with exponential decay of 0.97. Since we generate two reconstruction images for each input sample, we divide the reconstruction regularizer by half. During testing, we generate 1000 MultiMNIST images for each MNIST digit to have a fair comparison with the work by Sabour et al. \cite{sabour2017dynamic}, for a total of 10 million samples. We get a mean test error of $5.1\%_{\pm 0.005}$ with our model of 154K parameters, in comparison to the original work test error of $5.2\%$ with more than 9M parameters. Moreover, with an ensemble of the three models that get an accuracy greater than a threshold of 0.9470, we get a reduction of the test error to 3.8\%. These results show how our methodology is able to correctly detect and recognize highly overlapping digits encoding information about their position and style in the output layer capsules.

\subsection{Affine transformations embedding}
\begin{figure}[t]
    \centering
    \includegraphics[scale=0.45]{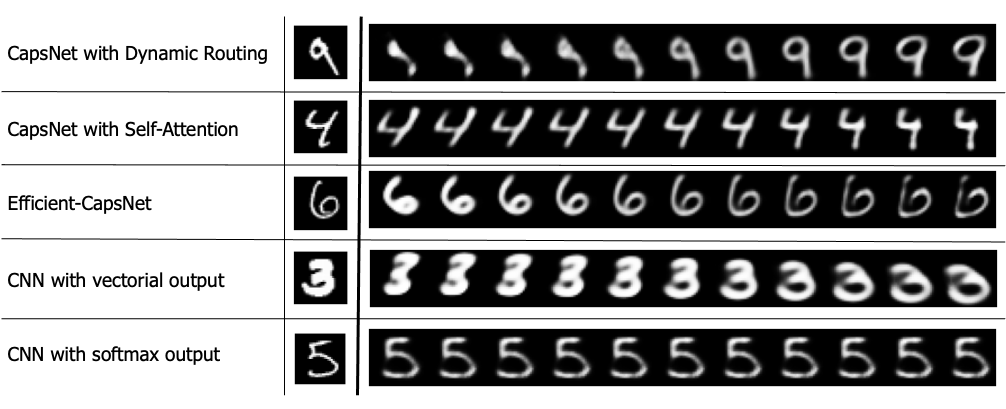}
    \caption{Effect on the digit reconstruction of the addition of perturbations to the output capsule values with different tested methodologies. All networks are able to embed shape, position and orientation information of the input digit except for the classical CNN with softmax output. That suggests that the capsule structure of the output, in which each class has its feature vector, is fundamental to get interpretable output embeddings.}
    \label{fig:affine_transformation}
\end{figure}

To understand what kind of information is embedded in the output capsules, we can perturb the prediction and observe how the reconstruction is affected. We select the capsule with the longest length and we add small positive and negative contributions to its single elements. Figure \ref{fig:affine_transformation} shows some example of perturbed images with different methodologies. We can observe how Efficient-CapsNet is behaving similarly to the original CapsNet \cite{sabour2017dynamic}, with the ability to encode combinations of different transformations of the digit. Retraining CapsNet also obtains similar behaviour with the proposed self-attention routing. A Convolutional Neural Network with a fake capsule layer, i.e. a vector instead of a scalar for each output class, also demonstrates the ability to encode actual shape, position and orientation information. On the other hand, considering the last features of a classical CNN, we are not able to reproduce this behaviour. That suggests that a capsule organization of the output, in which each digit has its instantiation parameters and the activation is measured by the length of the vector, is fundamental for a meaningful embedding of the information.
\begin{figure*}
\centering
\begin{subfigure}{0.45\textwidth}
  \centering
  \includegraphics[width=0.95\linewidth]{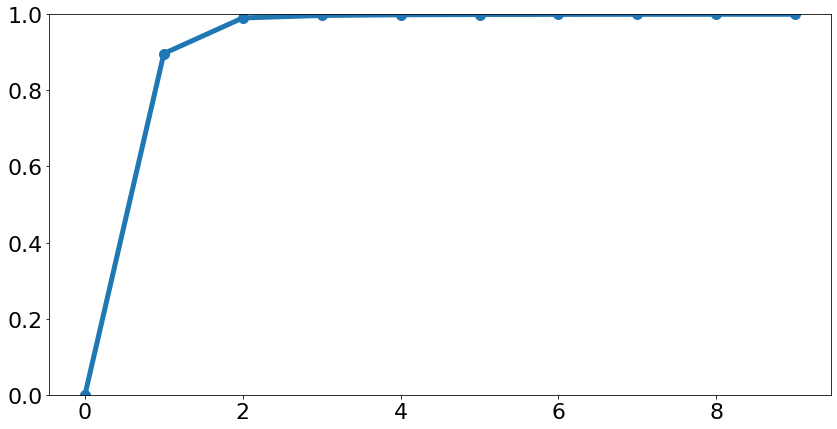}
  \caption{Translations on x: [-5,+5] pixels}
\end{subfigure}
\begin{subfigure}{0.45\textwidth}
  \centering
  \includegraphics[width=0.95\linewidth]{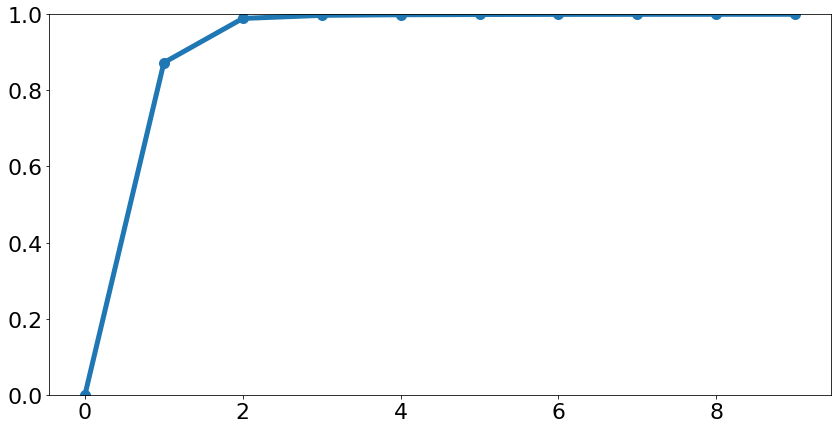}
  \caption{Translations on y: [-5,+5] pixels}
\end{subfigure}
\begin{subfigure}{0.45\textwidth}
  \centering
  \includegraphics[width=0.95\linewidth]{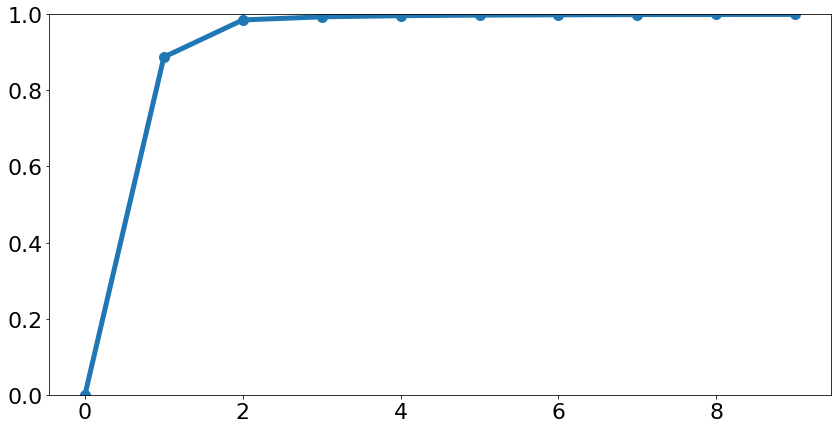}
  \caption{Rotations: [-25,+25] degrees}
\end{subfigure}
\begin{subfigure}{0.45\textwidth}
  \centering
  \includegraphics[width=0.95\linewidth]{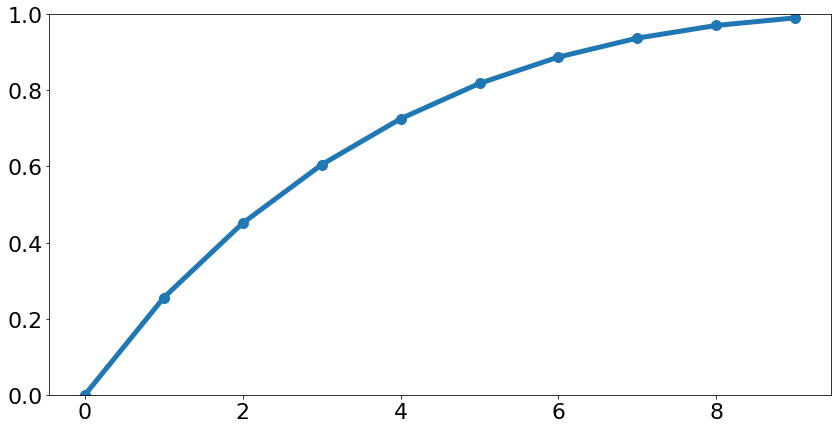}
  \caption{Random}
\end{subfigure}
\caption{Test set average cumulative variance explained with different numbers of PCA components by Efficent-CapsNet output capsule. It is clearly visible how the model is able to linearly embed affine transformations in the output space.}
\label{fig:PCA}
\end{figure*}

To further investigate the ability of the proposed model to capture meaningful information in the components of the output capsules, we study the equivariance to transformations with a method similar to the one proposed by Choi et al. \cite{choi2019attention}. For each test image we generate the images corresponding to the 11 translations between [-5,+5] pixels on both the axes and to the 51 rotations between [-25,+25] degrees. If the model is behaving as expected, we should see that each affine transformation (translation on x, translation on y, rotation) is independently linearly encoded in the activations of the correct output capsule. We verify it, by computing the Principal Component Analysis on the output vectors for each type of transformation.  We denote as $K$ the number of transformed images and with $N$ the number of output classes and we collect the output predictions $\textbf{\textit{u}}_i,\;i=1,...,K$. We center the data points and we compute the Singular Value Decomposition on the covariance matrix $C$:

\begin{ceqn}
\begin{gather}
    \textbf{\textit{z}}_i = \textbf{\textit{u}}_i - \overline{\textbf{\textit{u}}} \\
    \textbf{\textit{C}} = \frac{1}{K}\sum_{i=1}^K\textbf{\textit{z}}_i\,\textbf{\textit{z}}_i^T \\
    \textbf{\textit{C}} = \textbf{\textit{U}}\bm{\varSigma}\textbf{\textit{U}}^T
    \label{eq:PCA}
\end{gather}
\end{ceqn}

As a linearity metric, we consider the fraction of the first eigenvalue $\sigma_1$ of the matrix $\bm{\varSigma}$ over the sum of all its eigenvalues. Since the eigenvalues represent the variance of the original data points explained by each component of the PCA, if the transformations are linearly encoded, we should have a high fraction of the variance captured with just a single component, thus a high first eigenvalue ratio.

\begin{ceqn}
\begin{gather}
    r = \frac{\sigma_1}{\sum_{j=1}^N \sigma_j}
    \label{eq:PCA_ratio}
\end{gather}
\end{ceqn}

We perform this analysis on both the original CapsNet \cite{sabour2017dynamic} and our model. The average results on all the test images are shown in Table \ref{tab:PCA}, along with a comparison with the PCA performed on randomly generated vectors with the same dimension. Efficient-CapsNet shows higher linearity with respect to the original CapsNet in the encoding of affine transformations in the output capsule space. Figure \ref{fig:PCA} presents the average cumulative variance explained increasing the number of PCA components on the whole test set. For all the three transformations, Effienct-CapsNet is able to capture all the information with just two components, showing an almost perfectly linear behaviour with respect to the random example. That shows how our architecture can correctly embed position and orientation information of the recognized digit in the output vector components.

\begin{table}[h]
\centering
\begin{tabular}{lccc}
\toprule
Method    & Translations on x   & Translations on y   & Rotations   \\ \hline
Random                           & $25.57\%_{\pm0.028}$   & $25.54\%_{\pm0.028}$             & $13.49\%_{\pm0.009}$  \\ 
CapsNet \cite{sabour2017dynamic} & $83.78\%_{\pm0.006}$        & $79.82\%_{\pm0.009}$              & $88.01\%_{\pm0.006}$  \\ 
Efficient-CapsNet                & $89.69\%_{\pm0.005}$        & $87.28\%_{\pm0.008}$              & $88.75\%_{\pm0.005}$  \\ \bottomrule
\end{tabular}
\caption{Average percentage of variance captured by the first component of PCA performed on the output capsule vectors of the different transformations applied to test set images.}
\label{tab:PCA}
\end{table}

\section{Conclusion}
In this paper, we proposed Efficient-CapsNet, a novel capsule-based network that strongly highlights the generalization capabilities of capsules over traditional CNN, showing a much stronger knowledge representation after training. Indeed, our implementation, even with a very limited number of parameters is still capable of achieving state-of-the-art results on three distinct datasets, considerably outperforming previous implementations in terms of needed operations. Moreover, we introduced an alternative non-iterative routing algorithm that exploits a self-attention mechanism to rout a reduced number of capsules between subsequent layers efficiently. Further works will aim at designing a synthetic dataset to scale the network and analyze in-depth viewpoint generalization and network inner feature representations.

\section*{Acknowledgements}
This work has been developed with the contribution of the Politecnico di Torino Interdepartmental Centre for Service Robotics PIC4SeR\footnote{https://pic4ser.polito.it} and SmartData@Polito\footnote{https://smartdata.polito.it}.

\section*{Author contributions statement}
Conceptualization, V.M. and F.S.; methodology, V.M.; software, V.M. and F.S.; validation, V.M. and F.S.; formal analysis, V.M. and F.S.; investigation, V.M. and F.S.; resources, M.C.; data curation, V.M. and F.S.; writing original draft preparation V.M. and F.S.; writing review and editing, V.M. and F.S.; visualization, V.M. and F.S.; supervision, V.M. and F.S.; project administration, V.M., F.S. and M.C.; funding acquisition, M.C.

\bibliographystyle{unsrt}  
\bibliography{references}  


\end{document}